\documentclass[10pt,twocolumn,letterpaper]{article}

 \usepackage[pagenumbers]{cvpr} 

\definecolor{cvprblue}{rgb}{0.21,0.49,0.74}
\usepackage[pagebackref,breaklinks,colorlinks,allcolors=cvprblue]{hyperref}
\usepackage{mathtools}
\usepackage{subcaption} 
\usepackage{graphicx}
\usepackage{subcaption} 

\usepackage[ruled,vlined]{algorithm2e}

\def\paperID{10944} 
\def\confName{CVPR}
\def\confYear{2026}

\title{VeCoR — Velocity Contrastive Regularization for Flow Matching}

\author{
    Zong-Wei Hong \quad 
    Jing-lun Li \quad 
    Lin-Ze Li\textsuperscript{$\ddagger$} \quad 
    Shen Zhang \quad 
    Yao Tang\textsuperscript{$\dagger$} \\
    JIIOV Technology \\
    {\tt\small \{zongwei.hong, jinglun.li, linze.li, shen.zhang, yao.tang\}@jiiov.com}
}

\begin{document}
\maketitle
\makeatletter
\def\blfootnote{\gdef\@thefnmark{}\@footnotetext}
\makeatother
\blfootnote{\textsuperscript{$\dagger$} Corresponding author. \textsuperscript{$\ddagger$} Project leader.}
\begin{abstract}

Flow Matching (FM) has recently emerged as a principled and efficient alternative to diffusion models. Standard FM encourages the learned velocity field to follow a target direction; however, it may accumulate errors along the trajectory and drive samples off the data manifold, leading to perceptual degradation, especially in lightweight or low-step configurations.

To enhance stability and generalization, we extend FM into a balanced attract–repel scheme that provides explicit guidance on both “where to go” and “where not to go.” To be formal, we propose \textbf{Velocity Contrastive Regularization (VeCoR)}, a complementary training scheme for flow-based generative modeling that augments the standard FM objective with contrastive, two-sided supervision. VeCoR not only aligns the predicted velocity with a stable reference direction (positive supervision) but also pushes it away from inconsistent, off-manifold directions (negative supervision). This contrastive formulation transforms FM from a purely attractive, one-sided objective into a two-sided training signal, regularizing trajectory evolution and improving perceptual fidelity across datasets and backbones.

On ImageNet-1K 256$\times$256, VeCoR yields 22\% and 35\% relative FID reductions on SiT-XL/2 and REPA-SiT-XL/2 backbones, respectively, and achieves further FID gains (32\% relative) on MS-COCO text-to-image generation, demonstrating consistent improvements in stability, convergence, and image quality, particularly in low-step and lightweight settings.  
\textbf{Project page please see} \href{https://p458732.github.io/VeCoR_Project_Page/}{here}.
\vspace{-14px}
\end{abstract}

\section{Introduction}
\label{sec:intro}

\begin{figure}[htbp]
    \centering
    \includegraphics[width=\linewidth]{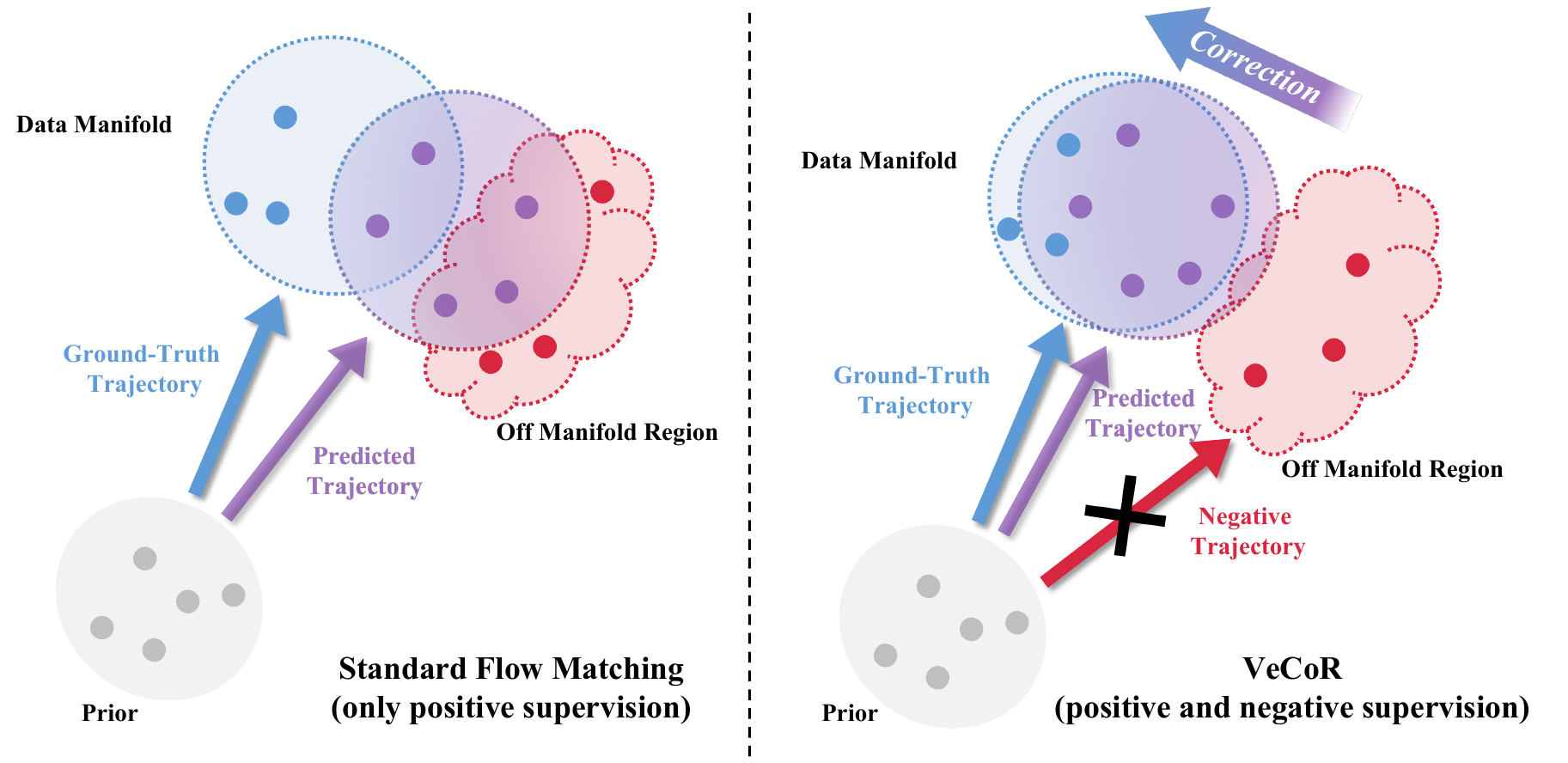}
    \caption{\textbf{Supervision and trajectory behavior.} \textbf{Left}—Standard Flow Matching (SFM): trained only with positive supervision toward the ground-truth velocity (blue), the predicted trajectory (purple) may slightly deviate from the data manifold, sometimes leading to less stable generations. \textbf{Right}—VeCoR: by contrastively suppressing negative trajectories (red path and ×), VeCoR adds negative supervision that discourages off-manifold deviations and guides trajectories back toward the data manifold, improving stability and perceptual fidelity.}
    \label{fig:teaser}
\vspace{-10px}
\end{figure}

\begin{figure*}[t]
\centering
\begin{subfigure}{0.2\textwidth}
    \centering
    \includegraphics[width=\linewidth]{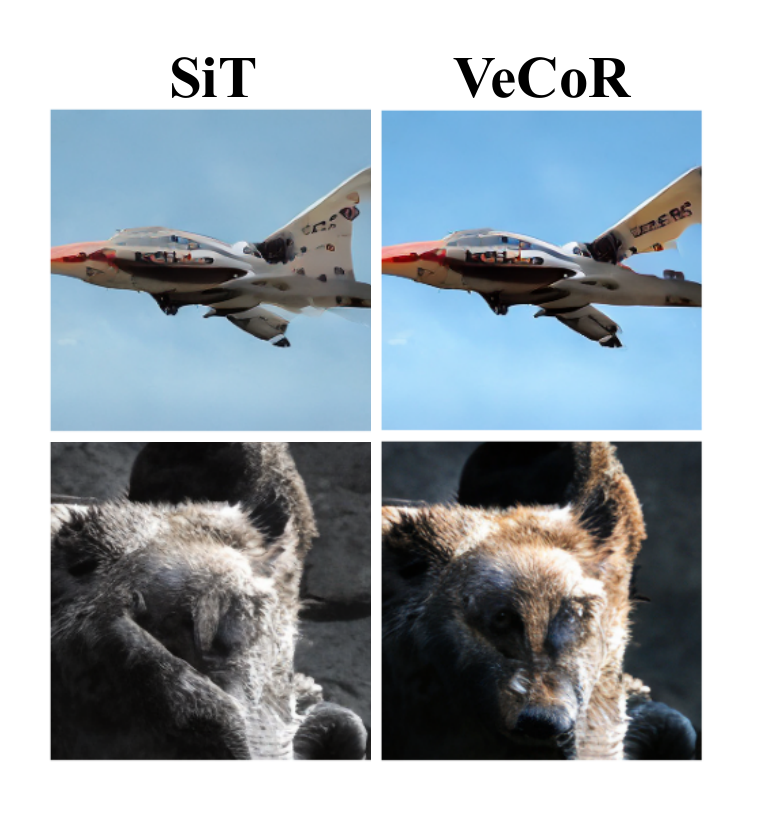}
    \caption{\textbf{Color/contrast}}
\end{subfigure}\hfill
\begin{subfigure}{0.2\textwidth}
    \centering
    \includegraphics[width=\linewidth]{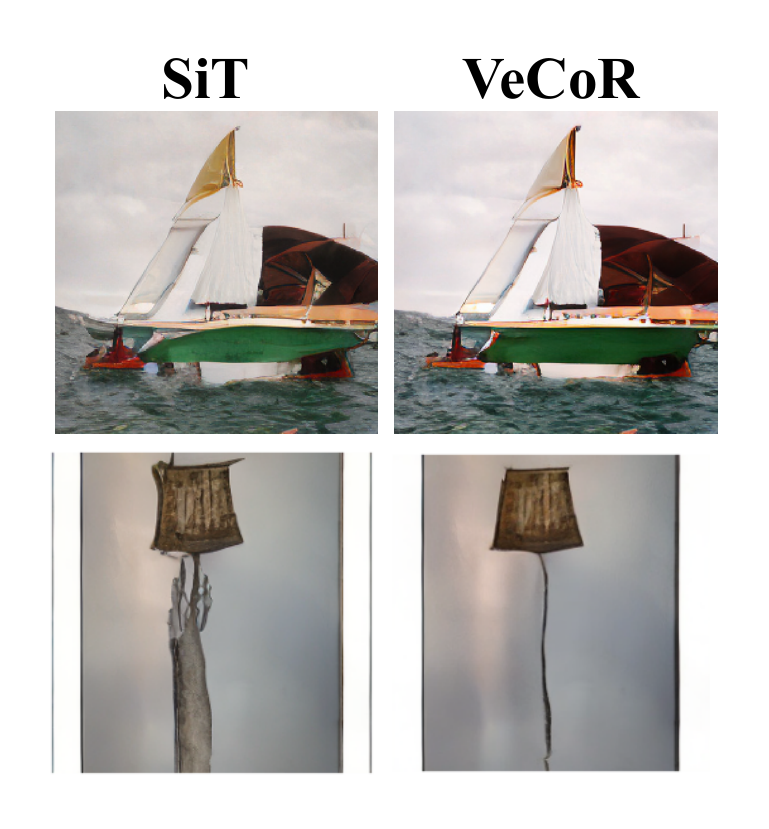}
    \caption{\textbf{Geometric consistency}}
\end{subfigure}\hfill
\begin{subfigure}{0.2\textwidth}
    \centering
    \includegraphics[width=\linewidth]{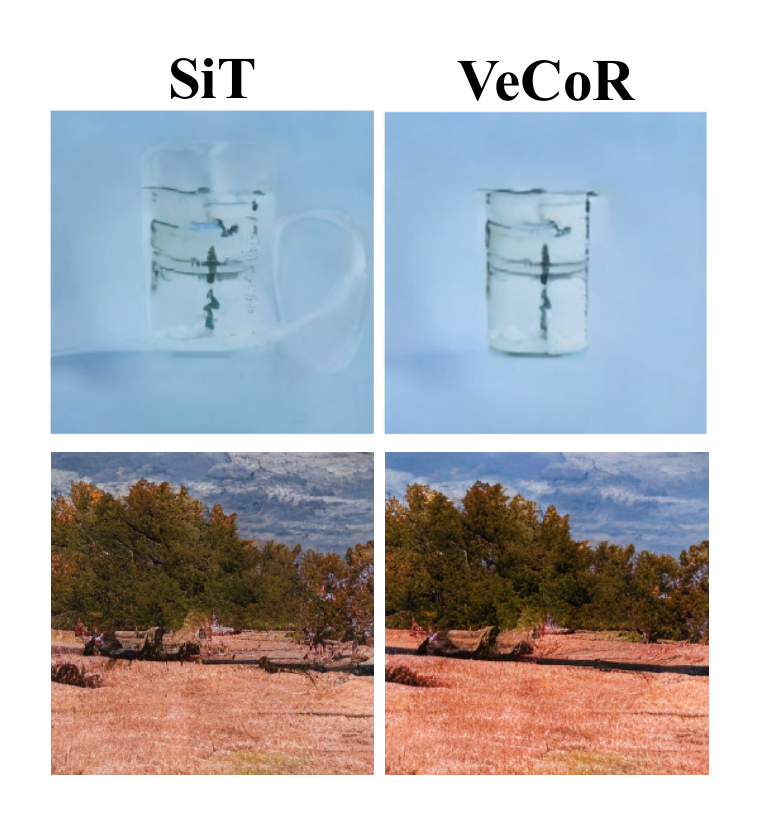}
    \caption{\textbf{Deblurring/sharpening}}
\end{subfigure}\hfill
\begin{subfigure}{0.2\textwidth}
    \centering
    \includegraphics[width=\linewidth]{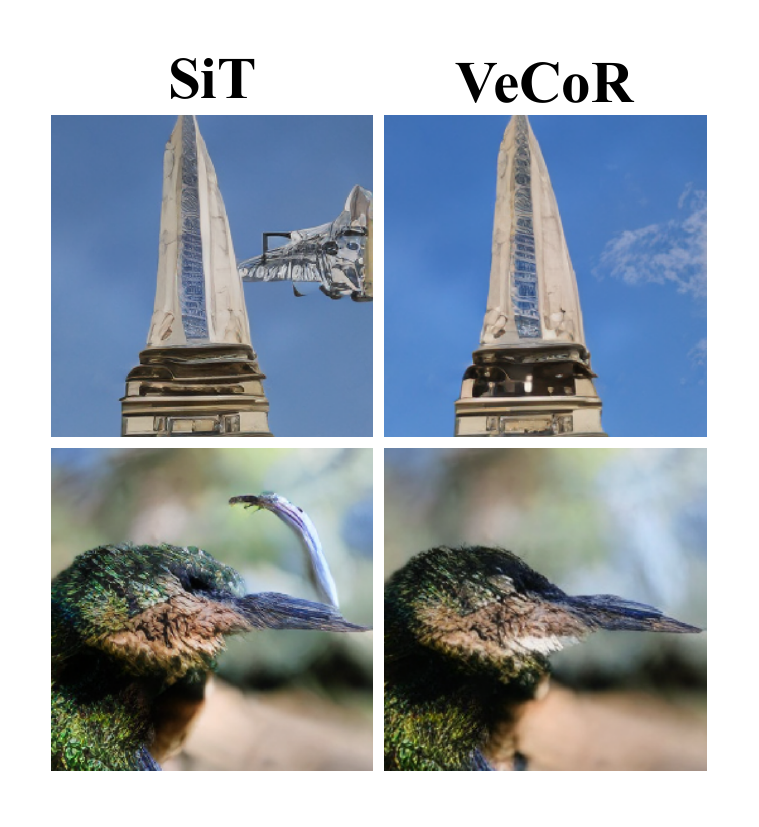}
    \caption{\textbf{Artifact removal}}
\end{subfigure}

\caption{\textbf{VeCoR refines strong SiT baselines by suppressing negative trajectories and improving stability and perceptual fidelity.}
Although SiT already produces plausible ImageNet-1K 256$\times$256 samples, its sampling trajectories can still drift from the ground truth, causing color/contrast shifts, geometric distortions, blur, and artifacts; VeCoR reduces these issues under identical sampling (same seed, 50 NFEs, Euler--Maruyama).
(a) \textbf{Color/contrast}: VeCoR yields a more saturated, uniform sky and wolf hues closer to the ground truth. 
(b) \textbf{Geometric consistency}: SiT bends the boat and distorts the lamp shade, while VeCoR produces a level hull and a lamp shade closer to the true shape. 
(c) \textbf{Deblurring/sharpening}: previously soft boundaries become crisp. 
(d) \textbf{Artifact removal}: SiT hallucinates extraneous structures (e.g., a mechanical arm near the spire; a protrusion above the bird’s beak), whereas VeCoR removes them, restoring clean, plausible shapes and textures.}

\label{fig:flowmatching_examples}
\vspace{-5px}
\end{figure*}

Flow Matching (FM)~\cite{lipman2022flow, liu2022rectifiedflow} learns a time-dependent velocity field that transports probability mass along a prescribed path between a reference distribution and the data. This viewpoint makes precise connections to diffusion/score-based modeling via the probability-flow formulation~\cite{song2021scorebasedgenerativemodelingstochastic, karras2022elucidating}, to continuous normalizing flows~\cite{chen2019neuralordinarydifferentialequations, grathwohl2018ffjordfreeformcontinuousdynamics}, and to optimal transport through dynamical formulations~\cite{benamou2000computational, peyre2019computational}. FM and its rectified variants have demonstrated competitive performance in image synthesis~\cite{lipman2022flow, liu2022rectifiedflow}.

While FM provides a theoretically elegant and empirically powerful foundation, subtle challenges can still arise in practice, particularly under lightweight or low-step configurations. In such settings, the integration process may accumulate minor inconsistencies in the learned velocity field, causing samples to drift slightly away from the data manifold, as illustrated in Fig.~\ref{fig:teaser} (left). This drift often manifests as mild perceptual degradations, such as desaturated colors, geometric misalignment, or blurred boundaries (see qualitative examples in Fig.~\ref{fig:flowmatching_examples}). These observations suggest that, although FM effectively directs samples toward the data manifold, it may benefit from complementary regularization that further stabilizes trajectory evolution and helps maintain perceptual consistency.

Building on extensive efforts to simplify and stabilize the transport process, including methods that enforce straighter ODE trajectories, reduce function evaluations, or leverage distillation techniques~\cite{salimans2022progressive, zhou2024simplefastdistillationdiffusion, lee2024improving, seong2025balancedconicrectifiedflow}, we revisit the supervision dynamics of Flow Matching from a complementary perspective. We move beyond a sole focus on the accuracy of predicted velocities and broaden the notion of supervision to encompass both attractive and repulsive guidance, yielding a more balanced treatment of trajectory learning. This perspective hypothesizes that incorporating a gentle repulsive component can further harmonize the learning dynamics, encouraging models not only to follow reliable flow directions but also to maintain stability and coherence along the manifold.

Building on the strong foundation of Flow Matching, we introduce \textbf{Velocity-Contrastive Regularization (VeCoR)}, a complementary training scheme designed to enhance the stability and robustness of learned velocity fields. VeCoR extends the conventional objective by jointly encouraging attraction toward ground-truth velocities and contrastive repulsion from dynamics-inconsistent counterparts. Rather than altering the core formulation of Flow Matching, VeCoR enriches it with supervision that goes beyond simple pointwise alignment by introducing \emph{negative velocity samples}---plausible yet gently perturbed directions that provide explicit contrastive cues for more balanced, two-sided guidance. These negatives are synthesized through semantic-preserving, augmentation-like perturbations applied across image, latent, and velocity domains, ensuring scalability, generality, and seamless integration into existing frameworks. This attract--repel formulation regularizes trajectory dynamics by suppressing drift along off-manifold directions and promoting correction toward the data manifold, as illustrated in the right panel of Fig.~\ref{fig:teaser}.


We empirically evaluate VeCoR across multiple image generation benchmarks and observe that it achieves higher sample quality and noticeably faster convergence than standard Flow Matching setups, while also improving training stability and generalization. The main contributions of this work are summarized as follows:
\begin{itemize}
\item We propose a complementary training scheme for flow-based generative models that augments standard supervision with an ensemble of stable and perturbed flows, improving sample quality and convergence without extra data or architectural changes.
\item We introduce Velocity Contrastive Regularization (VeCoR), a contrastive loss on the velocity field that enforces directional consistency of generative trajectories, yielding more stable and faster training.
\item Empirically, VeCoR delivers strong gains: on ImageNet-1K, it yields 22\% and 35\% relative FID reductions on SiT-XL/2 and REPA-SiT-XL/2 backbones, respectively, and a further 32\% FID reduction on MS-COCO text-to-image generation, indicating consistent improvements in stability, convergence, and image quality, especially in low-step and lightweight settings.

\end{itemize}
\section{Related Work}

Recent advances in generative modeling have been shaped by two dominant paradigms: diffusion-based models~\cite{ho2020denoising, song2020denoising, song2021scorebasedgenerativemodelingstochastic} and flow-matching (FM) approaches~\cite{lipman2022flow, liu2022rectifiedflow}.
Diffusion models formulate data generation as a stochastic denoising process that gradually perturbs data with noise and learns to reverse this process through score or noise prediction.
Extensive research has improved their stability and sampling efficiency through enhanced ODE/SDE solvers~\cite{song2020denoising, lu2022dpm, lu2025dpm, zheng2023dpm} and step-reduction or distillation techniques~\cite{salimans2022progressive, zhou2024simplefastdistillationdiffusion, meng2023distillationguideddiffusionmodels, feng2024relationaldiffusiondistillationefficient}.

Flow Matching (FM), by contrast, learns a continuous velocity field that deterministically transports a simple prior toward the data manifold~\cite{lipman2022flow, liu2022rectifiedflow}.
It offers a unified perspective connecting diffusion models, optimal transport~\cite{figalli2021invitation, peyre2019computational}, and continuous normalizing flows~\cite{grathwohl2018ffjordfreeformcontinuousdynamics, NEURIPS2018_69386f6b}, achieving diffusion-level generative quality with far fewer integration steps.
FM thus combines theoretical clarity with computational efficiency, motivating a growing body of work exploring its dynamics and extensions.

Building on these foundations, Stochastic Interpolants~\cite{albergo2025stochasticinterpolantsunifyingframework, liu2022rectifiedflow} further unify score-driven and velocity-driven formulations within a shared drift–diffusion structure, providing geometric insight into generative dynamics. Subsequent studies have sought to simplify or regularize the learned transport map by straightening trajectories, adopting adaptive integration schemes, or employing rectified formulations, thereby yielding smoother and more stable flows~\cite{lee2024improving, seong2025balancedconicrectifiedflow}. Together, these developments have progressively refined how continuous flows are represented and optimized.

While recent progress has significantly enhanced efficiency and smoothness, most objectives in FM remain directionally one-sided: they attract the model toward correct velocities but provide limited feedback on how to actively repel unstable or inconsistent dynamics. This suggests an opportunity for complementary training signals that jointly shape attractive and repulsive components of the flow.

A concurrent effort, Contrastive Flow Matching ($\Delta$FM)~\cite{stoica2025contrastive}, augments the FM objective with contrastive signals to enhance semantic discriminability. By pushing samples away from the data-averaged expectation, $\Delta$FM effectively reduces ambiguity between distinct conditions. In contrast, our approach, VeCoR, targets geometric stability within the vector field itself. Rather than focusing on inter-class separation, VeCoR applies contrastive regularization to rectify accumulated integration inconsistencies. By actively repelling the dynamics from corruption-induced drift, VeCoR tightens individual trajectories toward the true data manifold, ensuring structural coherence and robustness throughout the integration process.

\begin{figure*}[htbp]
    \centering
    \includegraphics[width=0.90\linewidth]{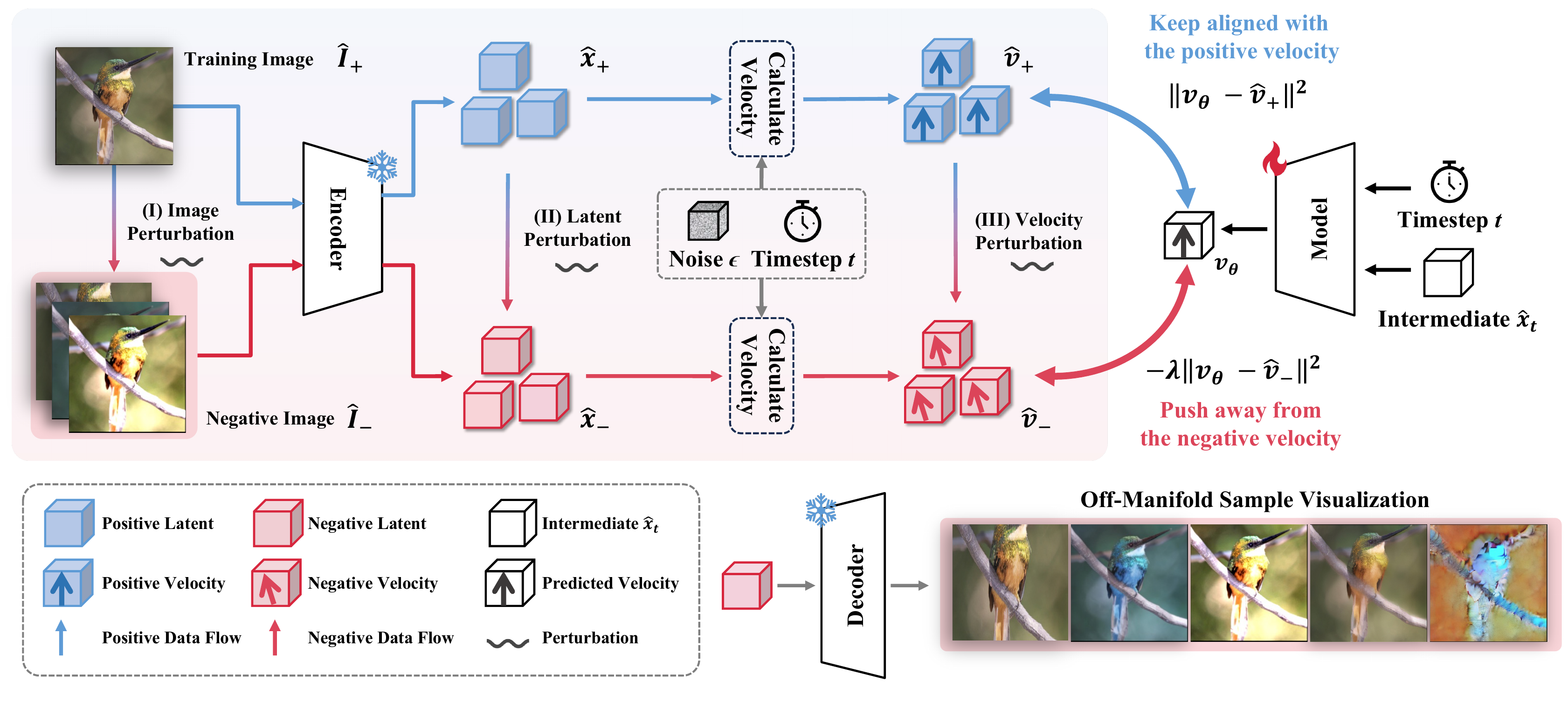}
\caption{
\textbf{Overview of the proposed Velocity-Contrastive Regularization (VeCoR) framework.}
VeCoR enhances flow matching (FM) by introducing a balanced, bidirectional supervision mechanism in the velocity space.
Instead of relying solely on positive guidance toward the ground-truth flow, VeCoR incorporates complementary contrastive cues that define counter-directional references across multiple representational domains.
These perturbations—spanning (I) image, (II) latent, and (III) velocity spaces—are implemented through lightweight, augmentation-like transformations that preserve semantic consistency while altering dynamic behaviors.
The resulting positive and negative velocities, $\hat{v}_{+}$ and $\hat{v}_{-}$, jointly guide the model-predicted velocity $v_\theta$ toward stable and coherent dynamics while discouraging drifts toward unstable regions.
The visualization (bottom right) illustrates how negative velocity guidance can induce off-manifold deviations, leading to degraded sample quality.
}
    \label{fig:framework}
\vspace{-5px}
\end{figure*}

\section{Preliminaries}
Building upon the problem formulation discussed in the introduction, this section formalizes the flow-matching process and establishes the mathematical groundwork that motivates our later velocity contrastive regularization.

\textbf{Problem Setup.} 
Given two arbitrary probability distributions, a prior $p_0$ and a target $p_1$, the objective of flow matching is to learn a vector field that transports samples from the former to the latter. In the context of generative modeling, $p_0$ is typically chosen as a simple distribution, such as a standard Gaussian $\mathcal{N}(0, I)$, while $p_1$ represents the true data distribution $p(x)$. Following the framework of Stable Diffusion~\cite{rombach2022highresolutionimagesynthesislatent}, we model the image distribution $p(x)$ in the latent space rather than pixel space. Specifically, each image $\hat{I}$ is first encoded into a latent representation $\hat{x}$ using a pretrained variational autoencoder encoder, upon which the flow field is learned to approximate the generative process.

\textbf{Stochastic Interpolants.}
For any given sample $\epsilon \sim \mathcal{N}(0, I)$ and data point $\hat{x} \sim p(x)$, flow matching progressively transforms noise into data over a continuous time interval. This transformation can be formalized as a time-dependent stochastic process using stochastic interpolants~\cite{albergo2025stochasticinterpolantsunifyingframework, ma2024sitexploringflowdiffusionbased}, defined by
\begin{equation}
    \label{eq:interpolation}
    \hat{x}_t = \alpha_t \hat{x} + \sigma_t \epsilon,
\end{equation}
where $\alpha_t$ and $\sigma_t$ denote time-dependent scheduling functions for $t \in [0, 1]$,
subject to the boundary conditions $\alpha_1 = \sigma_0 = 1$ and $\alpha_0 = \sigma_1 = 0$.
Although non-linear parameterizations of $\alpha_t$ and $\sigma_t$ are possible,
linear schedules are generally sufficient to achieve strong performance in diffusion models.
Therefore, in our experiments, we simply set $\alpha_t = t$ and $\sigma_t = 1 - t$.

\textbf{Learning Objective.}
The path induced by the interpolant in Eq.~\eqref{eq:interpolation} corresponds to the solution of a probability flow ordinary differential equation~\cite{song2021scorebasedgenerativemodelingstochastic, ma2024sitexploringflowdiffusionbased}, governed by a time-dependent velocity field $\hat{v}(\hat{x}, t, \epsilon)$. The model learns a neural network $v_{\boldsymbol{\theta}}(\hat{x}_t, t)$ that approximates this field. The ground-truth target velocity for the interpolated path is
\begin{equation}
\label{eq:velocity gt}
\hat{v} (\hat{x}, t, \epsilon) = \dot{\alpha}_t \hat{x} + \dot{\sigma}_t \epsilon,
\end{equation}
where $\dot{\alpha}_t$ and $\dot{\sigma}_t$ are time derivatives of the scheduling functions.

The standard flow-matching objective minimizes the mean-squared error (MSE) to this target:
\begin{equation}
\label{eq:org_fm_loss}
\mathcal{L}^{(\mathrm{FM})}(\theta)
= \mathbb{E}_{t, \hat{x}, \epsilon}\big[\|v_{\theta}(\hat{x}_t, t) - \hat{v} (\hat{x}, t, \epsilon)\|^2\big].
\end{equation}
With a finite training dataset 
$\mathcal{S}_{train}=\{(\hat{x}^{(i)}, t^{(i)}, \epsilon^{(i)})\}_{i=1}^N$,
this objective is implemented empirically as

\begin{equation}
\label{eq:empirical_fm_loss}
\begin{aligned}
\widehat{\mathcal{L}}^{(\mathrm{FM})}(\theta; \mathcal{S}_{train})
&= \frac{1}{N}\sum_{i=1}^{N}
\big\|v_{\theta}(\hat{x}^{(i)}_{t^{(i)}}, t^{(i)})  \\
&\quad - \hat{v}(\hat{x}^{(i)}, t^{(i)}, \epsilon^{(i)})\big\|^2.
\end{aligned}
\end{equation}

\textbf{Toward Better Regularization.}
While the empirical objective in Eq.~\eqref{eq:empirical_fm_loss} has proven highly effective for learning the overall transport dynamics, it primarily supervises where trajectories should move, offering limited guidance on where they should not.
Under constrained data or model capacity, this directional asymmetry may leave certain regions of the learned flow insufficiently regularized, allowing local instabilities to emerge.
These observations suggest an opportunity to enrich FM training with complementary signals that provide balanced guidance—not only encouraging accurate motion toward the data manifold, but also discouraging inconsistent or dynamically unstable directions within the velocity space.

\section{Method}
This section introduces \textbf{Velocity-Contrastive Regularization (VeCoR)}, a training scheme that augments standard Flow Matching with explicit negative guidance at the level of the velocity field.
Our key insight is to treat the learned velocity itself as editable data and to synthesize \emph{local negative velocity candidates} via augmentation-like perturbations in image, latent, or velocity space.
These negatives are semantically consistent but dynamically perturbed, and are used to repel the model away from unstable or off-manifold directions, while the standard FM loss continues to attract it toward the ground-truth flow.

We formalize the VeCoR objective in Sec.~\ref{sec:vecor} and describe negative velocity candidates in Sec.~\ref{sec: Negative Velocities Candidate Set}.

\subsection{Velocity-Contrastive Regularization}
\label{sec:vecor}

To enrich the supervision of flow matching, we introduce a contrastive regularization term that provides \emph{negative guidance}. Instead of solely aligning predicted and target velocities, VeCoR expands training into a two-sided process that attracts the model toward reliable flow directions while gently repelling it from unstable or off-manifold ones. This contrastive formulation complements the empirical FM objective by regularizing regions of the state space that remain underconstrained under finite data and model capacity, thereby improving flow stability and generative fidelity.

We view the predicted velocity field as editable data, from which informative negative samples can be synthesized. Concretely, we expand the finite training set by introducing semantically consistent yet dynamically perturbed velocity directions drawn from a pool of negative candidates (see Sec.~\ref{sec: Negative Velocities Candidate Set}). This augmentation-like perturbation scheme transforms each supervised instance into a set of semantically consistent yet statistically perturbed alternatives: one positive and several repulsive negatives, encouraging the model to refine its generative flow through explicit contrastive regularization.

Formally, for each velocity $\hat{v}_{+}^{(i)} \coloneqq \hat{v}(\hat{x}^{(i)}, t^{(i)}, \epsilon^{(i)}) \in \mathcal{S}_{\text{train}}$, we construct a finite set of candidate velocities:
\begin{equation}
\label{eq:vecor_cand_set_en}
\mathcal{C}_i \;=\; \big\{ \hat{ v}^{(i1)}_{-}, \ldots, \hat{ v}^{(iK)}_{-}\big\},
\end{equation}
where $K \in \mathbb{N}^{+}$ denotes the number of negative candidates per instance, and the negatives $\{\hat{ v}_{-}^{(ij)}\}_{j=1}^K$ are plausible yet misleading velocity directions. We then augment the training data via

\begin{equation}
\label{eq:train_aug_union}
\widetilde{\mathcal{S}}_{\text{train}}
=
\mathcal{S}_{\text{train}}
\;\cup\;
\bigcup_{i=1}^{N}\mathcal{C}_i .
\end{equation}
Given this augmented setup, learning proceeds by aligning with the positive and repelling from the negatives.
We regularize the model by pushing its predicted velocity away from the negative candidates while keeping it aligned with the true direction:
\begin{equation}
\label{eq:vecor_empirical}
\begin{aligned}
\widehat{\mathcal{L}}^{(\mathrm{VeCoR})}(\theta; \widetilde{\mathcal{S}}_{\text{train}})
&= \frac{1}{N}\sum_{i=1}^{N} \Big[
\big\|v_{\theta}(\hat{x}^{(i)}_{t^{(i)}}, t^{(i)})
-\hat{v}^{(i)}_{+}\big\|_2^{2} \\
&\quad - \lambda\sum_{j=1}^{K}
\big\|v_{\theta}(\hat{x}^{(i)}_{t^{(i)}}, t^{(i)})
-\hat{v}^{(ij)}_{-}\big\|_2^{2}
\Big].
\end{aligned}
\end{equation}

Here, $ \lambda \in (0, 1)$ controls the strength of the contrastive repulsion.

\subsection{Negative Velocity Candidate Set}
\label{sec: Negative Velocities Candidate Set}
Training FM with VeCoR requires plausible but incorrect velocity samples—those that appear semantically valid yet violate the underlying flow dynamics.
Instead of mining such samples from real-world data (which is costly and ill-defined), we leverage augmentation-like perturbations as a controllable and systematic perturbation mechanism. In the spirit of data augmentation commonly used in both supervised and unsupervised representation learning~\cite{bachman2019learning,henaff2020data,krizhevsky2009learning}, these perturbations naturally expose model fragilities while preserving semantic consistency.
This makes it suitable for constructing a scalable and diverse pool of negative velocity samples, consistent with the failure examples shown earlier in Fig.~\ref{fig:flowmatching_examples}.

Our perturbation pipeline follows the taxonomy introduced by SimCLR~\cite{chen2020simpleframeworkcontrastivelearning}, which broadly categorizes transformations into two types.
The first comprises spatial or geometric transformations, such as random cropping and resizing, channel shuffling, and CutMix~\cite{yun2019cutmixregularizationstrategytrain}.
The second includes appearance transformations, such as color jittering, Gaussian blur, and additive Gaussian noise.
Together, these complementary operations introduce controlled variations in both structure and appearance while preserving the underlying semantics of the data.

While conventional augmentation operates solely in
the image space, we reinterpret these operations as
augmentation-like perturbations and extend them to representational domains: image, latent, and velocity. Perturbations in each domain act at a distinct level of abstraction and are used to construct negative velocities that provide contrastive supervision, offering complementary perspectives on model robustness and feature alignment.

As illustrated in Fig.~\ref{fig:framework}, the process begins with a training image $\hat{I}_{+}$ and its perturbed counterpart $\hat{I}_{-}$, generated by image-level augmentation.
Passing them through the encoder yields latent representations $\hat{x}_{+}$ and $\hat{x}_{-}$.
Alternatively, latent-level augmentation can be directly applied to $\hat{x}_{+}$ to obtain a perturbed latent $\hat{x}_{-}$.
For simplicity, we use the unified notation $\hat{x}_{-}$ for both cases, as they serve the same role in constructing negative supervision.

Given a sampled noise $\epsilon$ and timestep $t$, we compute the corresponding positive and negative velocities, $\hat{v}_{+}$ and $\hat{v}_{-}$, from $\hat{x}_{+}$ and $\hat{x}_{-}$, respectively.
Additionally, velocity-level augmentation can be applied directly to $\hat{v}_{+}$ to produce a perturbed velocity $\hat{v}_{-}$.
Although $\hat{v}_{-}$ obtained from the encoder and $\hat{v}_{-}$ obtained through direct augmentation stem from different mechanisms, both represent semantically plausible yet dynamically inconsistent flows.
Hence, we collectively denote them as $\hat{v}_{-}$ for unified contrastive supervision.

This design establishes a flexible and extensible framework for constructing negative candidate velocities across multiple representational domains.
While our current implementation primarily demonstrates augmentation-based perturbations, the framework itself is not limited to this setting.
In essence, it provides a general mechanism for generating and contrasting representations, enabling broader applications such as domain adaptation, consistency regularization, or adversarial robustness.
During training, the model’s predicted velocity $v_\theta(x_t, t)$ is encouraged to align with the positive velocity $\hat{v}_{+}$ while being repelled from the negative velocity $\hat{v}_{-}$, thereby enforcing bidirectional contrastive regularization within this unified framework.
For completeness, the full set of augmentation-like perturbations and implementation details are provided in the supplementary material.

\section{Experiments}
This section first introduces the experimental setup and then presents the corresponding results.

\subsection{Experimental Settings}
\textbf{Datasets and Implementation.}
We evaluate both class-conditional and text-to-image (T2I) generation tasks.
For class-conditional generation, experiments are conducted on ImageNet-1k~\cite{deng2009imagenet} at $256\times256$ resolution, following the preprocessing of ADM~\cite{dhariwal2021diffusion}. Each image is encoded using the Stable Diffusion VAE~\cite{rombach2022high} into a latent tensor $\mathbf{z}\in\mathbb{R}^{32\times32\times4}$. We train vanilla SiT~\cite{albergo2023buildingnormalizingflowsstochastic} models of various scales (S/2, B/2, L/2, XL/2) under identical hyperparameters, except for the application of our VeCoR module.

To further evaluate the generalizability of our method, we integrate REPresentation Alignment (REPA)~\cite{yu2024representation} with VeCoR and train REPA models (B/2, XL/2) on ImageNet-1k at 256$\times$256 resolution. REPA accelerates training and enhances the generative quality of conventional diffusion models by aligning their intermediate representations with pretrained vision encoders (e.g., DiNOv2~\cite{oquab2023dinov2}) through an auxiliary distillation loss.

For T2I generation, we follow the setup of U-ViT~\cite{bao2023all, yu2024representation} and train REPA-MMDiT~\cite{esser2024scaling} from scratch on the MS-COCO dataset~\cite{lin2014microsoft}. The model is trained for 150K iterations with a batch size of 256, a hidden dimension of 768, and a depth of 24. Text embeddings are derived from the CLIP~\cite{radford2021learning} text encoder.

\textbf{VeCoR Settings.}
By default, negatives are formed in \emph{velocity space} via random channel shuffling with $K=1$, and we use a fixed contrastive weight $\lambda=0.05$. Variants in latent/image space and other operations are in Sec.~\ref{sec: ablation study}.

\textbf{Metrics.}
For class-conditional experiments, we report five standard metrics computed on 50,000 generated samples: Fréchet Inception Distance (FID)~\cite{heusel2017gans}, Inception Score (IS)~\cite{salimans2016improved}, spatial FID (sFID)~\cite{nash2021generating}, Precision (Prec.), and Recall (Rec.)~\cite{kynkaanniemi2019improved}.
For T2I, we report FID over the entire validation set.
Unless otherwise specified, we use the SDE Euler–Maruyama sampler with $w_t=\sigma_t$ and set the number of function evaluations (NFE) to 50.

\subsection{Main Results}


\begin{figure*}[htbp]
    \centering
    \includegraphics[width=0.85\linewidth]{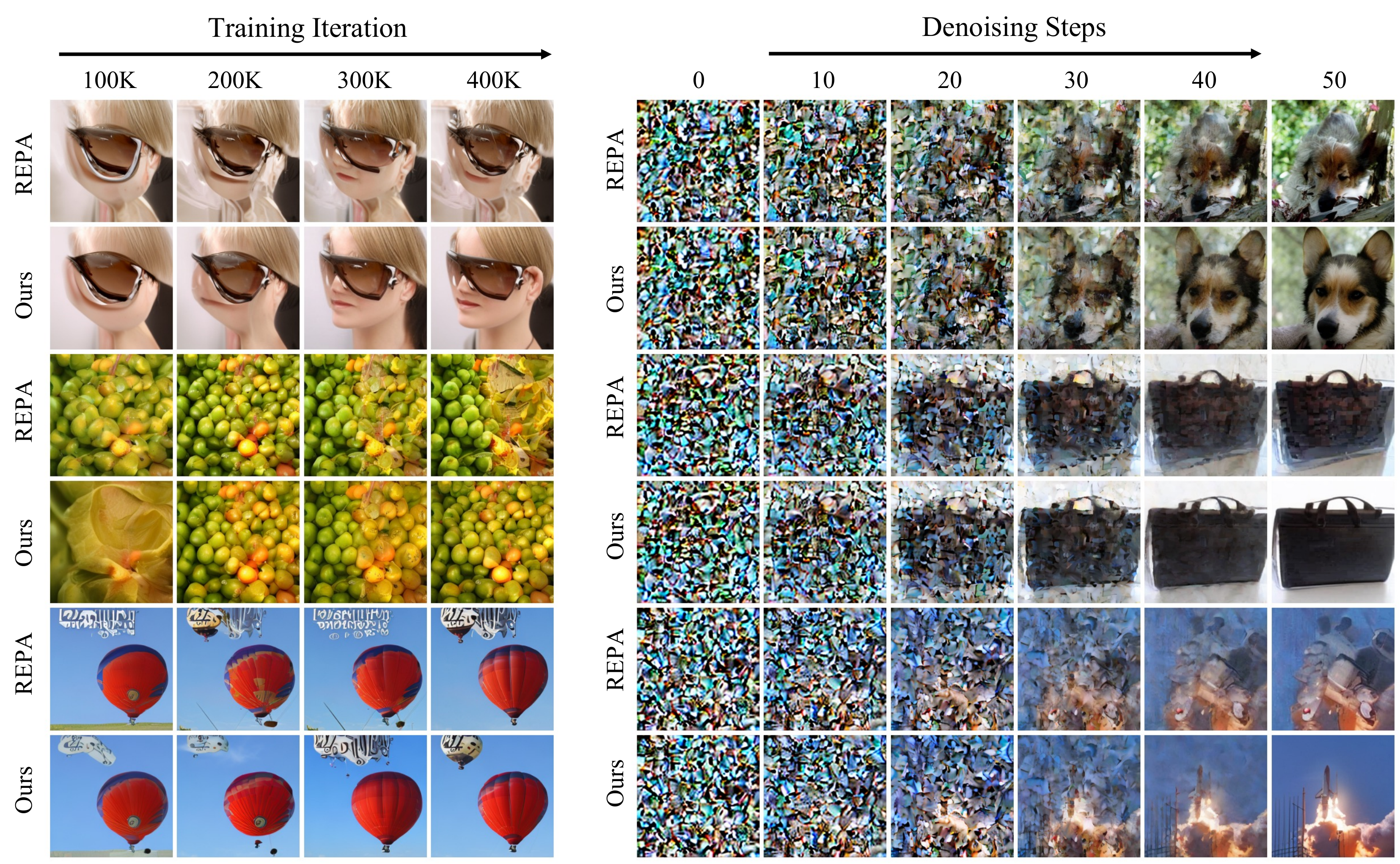}
    \caption{\textbf{Qualitative comparison between REPA and our REPA-based method (VeCoR) in terms of training convergence and denoising efficiency.}
    We compare the images generated by two SiT-XL/2 + REPA models during the first 400K iterations, one of which integrates our method, VeCoR. Both models share the same noise, sampler, and number of sampling steps, and neither uses classifier-free guidance. 
    The left panel shows results at different training iterations. While REPA demonstrates effectiveness in accelerating convergence, our VeCoR further improves the convergence speed. 
    The right panel illustrates the denoising process, showing that our method not only enhances training convergence but also enables the model to predict more reliable velocity fields and reconstruct the data manifold more accurately under low-step settings.}
    \label{fig:vis}
    \vspace{-10px}
\end{figure*}

\begin{table}[t]
    \centering
\caption{Main results on ImageNet-1K 256$\times$256 using SiT backbones (same seed, 50 NFEs, Euler--Maruyama), demonstrating the effectiveness of our method, VeCoR, across multiple model scales. ``--'' indicates results that do not exist in their paper.}

   \resizebox{0.8\linewidth}{!}{
\renewcommand{\arraystretch}{1.2}
\begin{tabular}{lccccc}
\toprule
\textbf{Model} & \textbf{FID}  $\downarrow$  & \textbf{IS}  $\uparrow$ & \textbf{sFID}  $\downarrow$  & \textbf{Prec.}  $\uparrow$ & \textbf{Rec.}  $\uparrow$ \\
\midrule

SiT-S/2~\cite{ma2024sitexploringflowdiffusionbased} & 64.26 & 22.78 & 15.60 & 0.39 & 0.56 \\
+\textit{$\Delta$}FM~\cite{stoica2025contrastive}& -- & -- & -- & -- & -- \\
+\textbf{VeCoR (Ours)} & \textbf{55.13} & \textbf{25.48} & \textbf{7.41} & \textbf{0.42} & \textbf{0.59} \\
\hline \addlinespace

SiT-B/2~\cite{ma2024sitexploringflowdiffusionbased} & 42.28 & 38.04 & 11.35 & 0.50 & 0.62 \\
+\textit{$\Delta$}FM~\cite{stoica2025contrastive} & 33.39 & 43.44 & \textbf{5.67} & 0.53 & \textbf{0.63} \\
+\textbf{VeCoR (Ours)} & \textbf{33.30} & \textbf{43.52} & 5.71 & \textbf{0.54} & \textbf{0.63} \\
\hline \addlinespace

SiT-L/2~\cite{ma2024sitexploringflowdiffusionbased} & 24.06 & 63.34 & 8.74 & 0.61 & 0.63 \\
+\textit{$\Delta$}FM~\cite{stoica2025contrastive} & -- & -- & -- & -- & -- \\
+\textbf{VeCoR (Ours)} & \textbf{18.86} & \textbf{71.12} & \textbf{4.92} & \textbf{0.64} & \textbf{0.62} \\
\hline \addlinespace

SiT-XL/2~\cite{ma2024sitexploringflowdiffusionbased} & 20.01 & 74.15 & 8.45 & 0.63 & 0.63 \\
+\textit{$\Delta$}FM~\cite{stoica2025contrastive} & 16.32 & 78.07 & 5.08 & 0.66 & \textbf{0.63} \\
+\textbf{VeCoR (Ours)} & \textbf{15.56} & \textbf{80.96} & \textbf{4.70} & \textbf{0.67} & 0.62 \\
\bottomrule
\end{tabular}}

\label{tab:imagenet-main-results}
\end{table}

\begin{table}[]
    \centering
\caption{Main results on ImageNet-1K 256$\times$256 using REPA-SiT backbones (same seed, 50 NFEs, Euler–Maruyama), demonstrating the generalization of our method.}
\resizebox{0.9\linewidth}{!}{
\renewcommand{\arraystretch}{1.2}
\begin{tabular}{lccccc}
\toprule
\textbf{Model} & \textbf{FID}  $\downarrow$ & \textbf{IS}  $\uparrow$ & \textbf{sFID}  $\downarrow$ & \textbf{Prec.}  $\uparrow$ & \textbf{Rec.}  $\uparrow$ \\
\midrule
REPA-SiT-B/2~\cite{yu2024representation} & 27.33 & 61.60 & 11.70 & 0.57 & \textbf{0.64} \\
\textbf{+VeCoR (Ours)} & \textbf{20.39} & \textbf{69.09} & \textbf{5.57} & \textbf{0.61} & \textbf{0.64} \\
\addlinespace
\hline
REPA-SiT-XL/2~\cite{yu2024representation} & 11.14 & 115.83 & 8.25 & 0.67 & \textbf{0.65} \\
\textbf{+VeCoR (Ours)} & \textbf{7.28} & \textbf{127.90} & \textbf{5.17} & \textbf{0.71} & 0.64 \\
\bottomrule
\end{tabular}}
\vspace{-10px}
\label{tab:repa-imagenet-results}
\end{table}





\begin{table}[t]
    \centering
    \caption{\textbf{Quantitative comparison on MS-COCO (Text-to-Image).} We report FID. \textbf{M+R} denotes the MMDiT+REPA baseline. Our method, \textbf{VeCoR}, is evaluated with two augmentation strategies: Random Crop and Resize (\textbf{RCR}) and Random Channel Shuffle (\textbf{RCS}).}
    \label{tab:repa-mmdit-results}
    \resizebox{0.90\linewidth}{!}{
        \begin{tabular}{l | c | c | c}
            \toprule
            \textbf{Solver} & \textbf{ODE (Heun)} & \textbf{SDE (E-M)} & \textbf{SDE (E-M)} \\
            \textbf{CFG Scale ($\omega$)} & 2.0 & 1.0 & 2.0 \\
            \textbf{Steps} & 50 & 50 & 50\\
            \midrule
            M + R (Reproduced) & 5.03 & 9.87 & 6.03\\
            + $\Delta$FM (Reproduced) & 5.16 & \textbf{6.64} & 4.78\\ 
            \midrule
            + \textbf{VeCoR} (w/ RCS) & 5.30 & 6.65 & 5.03 \\ 
            + \textbf{VeCoR} (w/ RCR) & \textbf{4.82} & 7.95 & \textbf{4.55} \\ 
            \bottomrule
        \end{tabular}
    }
\end{table}

\begin{table}[t]
    \centering
\caption{Quantitative results on ImageNet 256$\times$256 (NFE=50). We report the best results for each model after conducting a grid search for classifier-free guidance (CFG) hyperparameters over $w \in \{1.25, 1.75, 1.8, 1.85, 2.25\}$, $\sigma_{\text{low}} = 0$, and $\sigma_{\text{high}} \in \{0.50, 0.65, 0.75, 1.0\}$. }
    \label{tab:imagenet-results}
    \resizebox{0.75\linewidth}{!}{
        \begin{tabular}{l ccc cc}
            \toprule
            \raisebox{-1.5ex}{Method} & \multicolumn{3}{c}{CFG Terms} & \multicolumn{2}{c}{Metric} \\
            
            \cmidrule(lr){2-4} \cmidrule(lr){5-6}
            
             & $w$ & $\sigma_{\text{low}}$ & $\sigma_{\text{high}}$  & FID$\downarrow$ & sFID$\downarrow$ \\
            \midrule
            
            REPA SiT-XL/2 & 1.75 & 0.0 & 0.75 & 2.09 &  5.55 \\
            
            + $\Delta$FM & 1.85 & 0.0 & 0.65 & 1.97 & 4.49 \\
            + VeCoR & 1.85 & 0.0 & 0.65 & \textbf{1.94} & \textbf{4.45} \\
            \bottomrule
        \end{tabular}
    }
\vspace{-10px}
\end{table}
\textbf{Class-Conditional on ImageNet-1K}
Table~\ref{tab:imagenet-main-results} reports ImageNet-1K results with SiT backbones (S/B/L/XL). Compared to SiT baselines, VeCoR \emph{consistently} improves nearly all metrics—especially for smaller models (FID $\downarrow$14–22\%, sFID $\downarrow$44–53\%), while recall is largely preserved and only slightly reduced at L/2 and XL/2. This suggests that explicitly constraining off-manifold directions helps the model estimate more accurate velocities and recover finer spatial details. 
Against the contrastive baseline $\Delta$FM~\cite{stoica2025contrastive}, VeCoR is on par at B/2 and clearly stronger at XL/2 (lower FID/sFID with higher IS), indicating that the regularization benefits scale with model capacity.

\textbf{Train with REPA}
Table~\ref{tab:repa-imagenet-results} reports results with REPA-SiT backbones under a fixed 50-NFE budget at $256\times256$. Relative to the REPA-SiT baselines, VeCoR reduces FID by 25–35\% (27.33$\rightarrow$20.39 on B/2; 11.14$\rightarrow$7.28 on XL/2), improves sFID by 37–52\% (11.70$\rightarrow$5.57; 8.25$\rightarrow$5.17), increases IS (61.60$\rightarrow$69.09; 115.83$\rightarrow$127.90), and largely maintains Recall (flat at B/2 and only slightly reduced at XL/2), with small Precision gains. These results demonstrate the strong generalization ability of VeCoR across architectures and model scales.

\textbf{Text-to-image on MS-COCO}
\label{sec:t2i-coco}
We evaluate VeCoR on the MS-COCO dataset using the MMDiT+REPA pipeline. To provide a comprehensive comparison, we report results for both ODE (Heun, Steps=50) and SDE (Euler-Maruyama, Steps=50) solvers across different classifier-free guidance (CFG) scales. Table~\ref{tab:repa-mmdit-results} reports the FID scores.

Integrating VeCoR yields substantial improvements over the MMDiT+REPA baseline across all evaluated settings. Notably, under a higher guidance scale ($\omega=2.0$), VeCoR with \textit{Random Crop and Resize} (RCR) achieves the best overall performance, reaching an FID of \textbf{4.82} with the ODE solver and \textbf{4.55} with the SDE solver. In both cases, VeCoR significantly outperforms the $\Delta$FM baseline (5.16 and 4.82, respectively). Under lower guidance (SDE, $\omega=1.0$), VeCoR with \textit{Random Channel Shuffle} (RCS) drastically reduces the baseline FID from 9.87 to 6.65, achieving performance comparable to $\Delta$FM (6.64). Overall, VeCoR consistently enhances high-fidelity text-to-image generation on MS-COCO.

\textbf{Combining with Classifier-Free Guidance.}
To further push performance limits, we combine VeCoR with Classifier-Free Guidance (CFG)~\cite{ho2022classifier}. Similar to $\Delta$FM~\cite{stoica2025contrastive}, our contrastive training objective can be mathematically interpreted as steering the predicted velocities away from the \emph{mean of the synthesized off-manifold trajectories}. Consequently, naively applying CFG—which independently steers predictions away from the standard unconditional flow—can create conflicting guidance signals, leading to suboptimal generation quality. To mitigate this, we follow the strategy proposed in~\cite{stoica2025contrastive} to rectify this conflict by explicitly incorporating the pre-computed mean of these off-manifold trajectories into the modified guidance equation. 

Through a rigorous grid search over the CFG scale $w$ and interval~\cite{kynkaanniemi2024applying} $[\sigma_{\text{low}}, \sigma_{\text{high}}]$, VeCoR achieves an FID of \textbf{1.94} and an sFID of \textbf{4.45}. By surpassing $\Delta$FM (FID 1.97) under \textit{identical} optimal hyperparameters, VeCoR demonstrates a more robust vector field that effectively leverages guidance, establishing a new state-of-the-art for this architecture.

\textbf{Qualitative Results}
Qualitative comparisons between REPA-SiT-XL/2 and REPA-SiT-XL/2+VeCoR on ImageNet are presented in Fig.~\ref{fig:vis}, while additional text-to-image results are provided in the supplementary material.

\subsection{Ablation Study}
\textbf{Ablation Study on Negative Velocity Candidate Set}
\label{sec: ablation study}
As presented in Table~\ref{tab:ablation-perturbation-space}, several perturbation-based variants outperform the baseline SiT-S/2 (FID = 64.26), indicating that introducing structured perturbations across different representational domains can enhance generative quality. When grouped by augmentation type, \textit{spatial/geometric transformations} (e.g., random cropping, channel shuffling, and CutMix) generally achieve lower FID scores than \textit{appearance transformations} (e.g., color jitter, Gaussian blur, and noise), particularly in the \textit{velocity space}. This trend suggests that geometric perturbations, which primarily modify structural composition while preserving semantic integrity, generate more informative and dynamically consistent negative candidates. In contrast, appearance-based perturbations often introduce shallow visual variations that provide weaker supervision signals. Overall, these results highlight the advantage of modeling structural variability in the velocity domain to improve contrastive alignment and synthesis fidelity.

\textbf{Ablation on Negative Set Size and Operators.}
As shown in Table~\ref{tab:ablation-k-scale}, \textit{Channel Shuffle} with $K{=}2$ negative candidates achieves the optimal trade-off between fidelity and diversity. Increasing $K$ beyond 2 yields diminishing returns in FID but modestly improves recall, while $K{=}1$ underperforms due to insufficient regularization diversity. \textit{Random Crop} consistently lags behind channel-level perturbations and shows little sensitivity to $K$. Finally, combining operators (\textit{Shuffle + Crop}) introduces redundant variations, failing to improve upon \textit{Channel Shuffle} alone.

\begin{table}[t!]
\centering
\caption{
\textbf{FID comparison under different perturbation spaces (columns) and augmentation types (rows).}
Baseline FID (64.26) of SiT-S/2 is shown for reference. Lower is better.
}
\resizebox{0.9\linewidth}{!}{
\begin{tabular}{lccc}
\toprule
\textbf{Perturbation Type / Operation} & \textbf{Image} & \textbf{Latent} & \textbf{Velocity} \\
\midrule
\multicolumn{4}{l}{\textit{Baseline (SiT-S/2):} FID = 64.26} \\
\midrule
\addlinespace[3pt]
\multicolumn{4}{l}{\textbf{Spatial / Geometric Transformations}} \\
\addlinespace[3pt]

Random Channel Shuffle & 62.4 & \textbf{60.77} & \textbf{55.13} \\

Random Crop \& Resize  & 62.87 & 62.79 & 59.43 \\

CutMix                 & \textbf{61.94} & 62.98 & 59.62 \\
\addlinespace[5pt]
\multicolumn{4}{l}{\textbf{Appearance Transformations}} \\
\addlinespace[3pt]

Gaussian Blur           & 64.18 & 62.28 & 58.73 \\

Gaussian Noise          & 63.39 & 64.22 & 65.07 \\

Color Jitter            & 63.63 & 65.18 & 64.69 \\
\bottomrule
\end{tabular}}
\vspace{-10px}
\label{tab:ablation-perturbation-space}
\end{table}

\begin{figure}[t]
\centering
\begin{minipage}[t]{0.95\linewidth}
    \centering
    \begin{subfigure}[t]{\linewidth}
        \centering
        \includegraphics[width=1\linewidth]{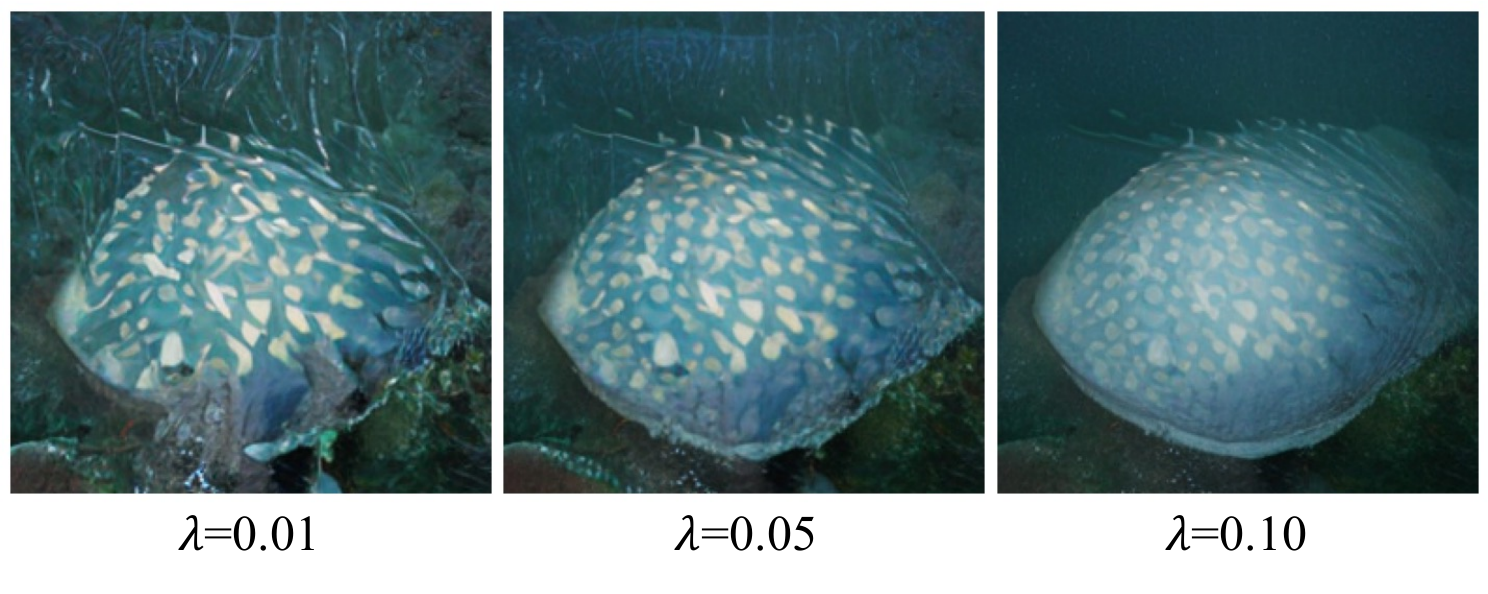}
        \caption{Visalization of different $\lambda$ settings}
        \label{fig:velocity_error}
    \end{subfigure}

    \vspace{-2px} 

    \begin{subfigure}[t]{\linewidth}
        \centering
        \includegraphics[width=0.5\linewidth]{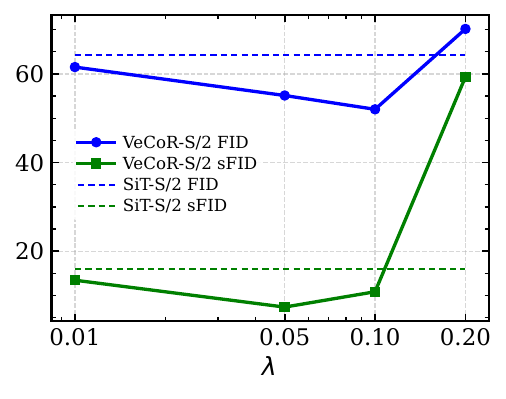}
        \caption{Qualitative analysis of the effect of regularization coefficient $\lambda$. }
        \label{fig:lambda_effect}
    \end{subfigure}
\vspace{-5px}
\caption{
\textbf{Ablation on the regularization coefficient $\lambda$.} 
Comparison illustrating that a moderate $\lambda$ (=0.05) yields the most natural and detailed images, while smaller or larger values cause artifacts or over-smoothed geometry.
}
    \label{fig:overall_lambda_vis}
\end{minipage}
\vspace{-5px}
\end{figure}

\textbf{Ablation on the effect of $\lambda$.}
The effect of the regularization coefficient $\lambda$ is illustrated in Fig.~\ref{fig:overall_lambda_vis}. 
When $\lambda$ is too small (e.g., 0.01), VeCoR slightly improves both FID and sFID compared to SiT, but the generated images still exhibit noticeable artifacts. 
A moderate value (around 0.05) achieves the best trade-off, yielding lower FID and sFID scores and producing visually sharper and more natural results. 
However, as $\lambda$ increases further (0.1–0.2), excessive regularization constrains the model’s generative capacity, leading to the loss of fine-grained details.

\begin{figure}[t]
  \centering
  \begin{subfigure}{0.49\linewidth}
    \centering
    \includegraphics[width=\linewidth]{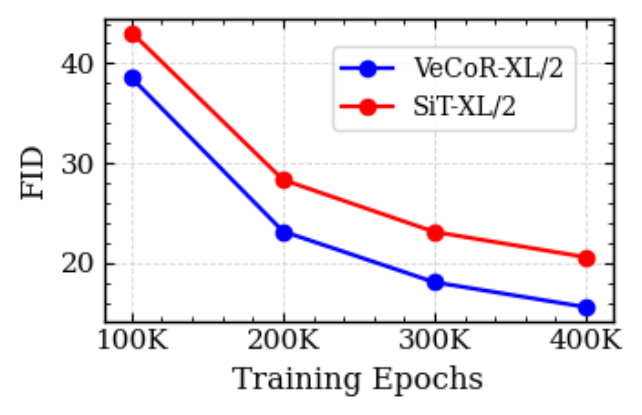}
    \caption{FID vs.\ Training Epochs}
    \label{fig:row-a}
  \end{subfigure}\hfill
  \begin{subfigure}{0.49\linewidth}
    \centering
    \includegraphics[width=\linewidth]{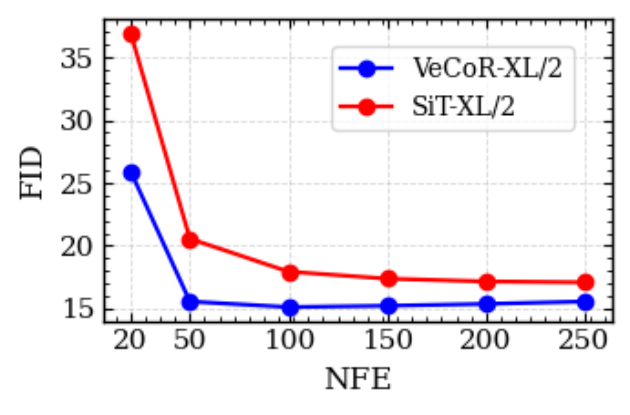}
    \caption{FID vs.\ NFE}
    \label{fig:row-b}
  \end{subfigure}
  \vspace{-7px}
\caption{\textbf{Training dynamics and sampling efficiency.} \textbf{(a)} SiT-XL/2+VeCoR (blue) converges faster and yields a lower FID than the baseline (red). \textbf{(b)} VeCoR attains lower FID at small NFE ($\le$50) and remains comparable at larger NFE.}

  \label{fig:two-in-row}
\vspace{-17px}
\end{figure}

\textbf{Training Dynamics and Sampling Efficiency.}
As shown in Fig.~\ref{fig:two-in-row}, we compare the models under varying training and sampling budgets. In (a), while both models improve over epochs, SiT-XL/2+VeCoR converges faster to a lower FID. In (b), FID drops sharply up to $\sim$50 NFE before saturating; notably, VeCoR achieves better FID in low-NFE settings while remaining competitive at higher NFE. Together, these trends highlight that VeCoR accelerates both training convergence and sampling efficiency without asymptotic degradation.

\begin{table}[t!]
\centering
\caption{
\textbf{Ablation on negative set size ($K$) and operators.} \textit{Channel Shuffle} ($K{=}2$) achieves the best FID/sFID and IS/precision. Larger $K$ marginally improves recall without fidelity gains. \textit{Random Crop} is inferior and $K$-insensitive. The combined \textit{Shuffle+Crop} underperforms \textit{Channel Shuffle} alone.
}
\resizebox{\linewidth}{!}{
\renewcommand{\arraystretch}{1.2}
\begin{tabular}{lccccc}
\toprule
\textbf{Method} & \textbf{FID} $\downarrow$ & \textbf{IS} $\uparrow$ & \textbf{sFID} $\downarrow$ & \textbf{Precision}  $\uparrow$  & \textbf{Recall}  $\uparrow$ \\
\midrule
Velocity (Random Crop, $K{=}1$) & 59.43 & 24.58 & 9.69 & 0.415 & 0.580 \\
Velocity (Random Crop, $K{=}2$) & 59.34 & 24.48 & 9.65 & 0.415 & 0.579 \\
Velocity (Channel Shuffle, $K{=}1$) & 55.13 & 25.48 & 7.406 & 0.416 & 0.592 \\
Velocity (Channel Shuffle, $K{=}2$) & \textbf{52.60} & \textbf{26.21} & \textbf{6.89} & \textbf{0.420} & 0.596 \\
Velocity (Channel Shuffle, $K{=}3$) & 53.96 & 25.47 & 7.00 & 0.416 & 0.596 \\
Velocity (Channel Shuffle, $K{=}4$) & 53.70 & 25.83 & 6.99 & 0.415 & \textbf{0.599} \\
Velocity (Shuffle + Crop, $K{=}2$) & 56.16 & 25.59 & 7.41 & 0.415 & 0.591 \\
\bottomrule
\end{tabular}}

\label{tab:ablation-k-scale}
\vspace{-1em}
\end{table}

\section{Conclusion}
We presented \textbf{Velocity Contrastive Regularization (VeCoR)}, a lightweight and general training scheme that extends flow matching beyond one-sided supervision, without any additional networks or external data.
By coupling attraction toward reliable velocity directions with contrastive repulsion from dynamically inconsistent ones, VeCoR provides balanced, two-sided guidance that stabilizes learning and accelerates convergence.
Across ImageNet and MS-COCO benchmarks, VeCoR improves fidelity and robustness, sharper structures, and faster training under the same computational budget.
While the current negative sampling strategy remains heuristic and data-agnostic, future work will explore adaptive hard-negative mining and trajectory-aware perturbations to strengthen stability and efficiency further.
Overall, VeCoR offers a straightforward, plug-and-play approach to more stable, data-efficient, and unified schemes for continuous generative modeling.

{
    \small
    \bibliographystyle{ieeenat_fullname}
    \bibliography{main}
}


\clearpage
\setcounter{page}{1}
\maketitlesupplementary
\appendix
\section{More Implementation Details}
This section elucidates more details about the concrete augmentation-like perturbation in~\cref{sec: Negative Velocities Candidate Set}.

\subsection{Augmentation-like Perturbation Details}

Given a batch input $z_{+} \in \mathbb{R}^{B \times C \times H \times W}$,
which may represent training images ($\hat{I}_{+}$), latents ($\hat{x}_{+}$), or velocities ($\hat{v}_{+}$), our goal is to apply augmentation-like perturbations to obtain negative samples $z_-$.
\paragraph{Random Channel Shuffle}
We apply a per-sample cyclic channel shift to ensure that no channel remains in its original position.  
Given
$
z_+ \in \mathbb{R}^{B \times C \times H \times W},
$
a shift
$
k \in \{1,\ldots,C-1\}
$
is sampled and applied via modular indexing, producing the perturbed output \( z_- \).

\paragraph{Random Crop and Resize}
For a batch input 
$
z_+ \in \mathbb{R}^{B \times C \times H \times W},
$
we uniformly sample the target area ratio and aspect ratio:
\[
\alpha \sim \mathcal{U}(\text{scale}_{\min},\, \text{scale}_{\max}), \qquad
r \sim \mathcal{U}(\text{ar}_{\min},\, \text{ar}_{\max}),
\]
with default \(\alpha \in [0.9, 0.95]\) and \(r \in [0.95,1.05]\).

The target crop area is
\[
A_{\text{crop}} = \alpha (HW),
\]
and the crop dimensions are
\[
h = \sqrt{\frac{A_{\text{crop}}}{r}}, \qquad
w = \sqrt{A_{\text{crop}}\, r},
\]
rounded to integers and clamped to valid ranges.  
If the sampled dimensions fall below a threshold, we fall back to a larger crop (e.g., \(0.9H \times 0.9W\)).  
A valid crop location is sampled uniformly, and the cropped region is resized back to \((H, W)\), resulting in \(z_-\).

\paragraph{CutMix}
Given a batch
$
z_{+} \in \mathbb{R}^{B \times C \times H \times W},
$
we first construct a derangement permutation 
\[
\pi : \{1,\ldots,B\} \rightarrow \{1,\ldots,B\}
\]
that satisfies \(\pi(i) \neq i\) for every index \(i\), ensuring that no sample is mixed with itself.

For each sample \(z_{+}^{(i)}\), we draw a mixing coefficient
\[
\lambda^{(i)} \sim \mathrm{Beta}(\alpha, \alpha), \qquad \alpha = 1,
\]
and compute the CutMix region scale
\[
r^{(i)} = \sqrt{1 - \lambda^{(i)}}.
\]
The corresponding box width and height are
\[
w^{(i)} = r^{(i)} W, 
\qquad
h^{(i)} = r^{(i)} H.
\]

A box center \((c_x, c_y)\) is sampled uniformly over the spatial domain.  
The bounding coordinates are then clipped to valid image ranges:
\[
x_1 = \mathrm{clip}\!\left(c_x - \frac{w^{(i)}}{2},\, 0,\, W\right),
\]
\[
x_2 = \mathrm{clip}\!\left(c_x + \frac{w^{(i)}}{2},\, 0,\, W\right),
\]
\[
y_1 = \mathrm{clip}\!\left(c_y - \frac{h^{(i)}}{2},\, 0,\, H\right),
\]
\[
y_2 = \mathrm{clip}\!\left(c_y + \frac{h^{(i)}}{2},\, 0,\, H\right).
\]

Finally, the rectangular region of \(z_{+}^{(i)}\) within 
\((x_1:x_2,\, y_1:y_2)\) is replaced by the corresponding patch from the paired sample \(z_{+}^{(\pi(i))}\), yielding the CutMix-perturbed output \(z_-^{(i)}\).

\paragraph{Gaussian Blur}
Given a batch input
$
z_+ \in \mathbb{R}^{B \times C \times H \times W},
$
we apply a per-channel Gaussian blur with kernel size \(k\) (odd) and standard deviation \(\sigma \ge 1\).
We use \(k=5\) and \(\sigma=1\).

The kernel is
\[
G(u,v) = \exp\!\left(-\frac{u^{2} + v^{2}}{2\sigma^{2}}\right),
\qquad
u,v \in \bigl[-\tfrac{k-1}{2},\, \tfrac{k-1}{2}\bigr],
\]
normalized so that \(\sum_{u,v}G(u,v)=1\).

We replicate the kernel across channels:
\[
K \in \mathbb{R}^{C \times 1 \times k \times k}, \qquad K_c = G,
\]
and apply depthwise convolution with reflection padding
\[
p = \left\lfloor \frac{k}{2} \right\rfloor,
\]
which prevents artificial dark borders or edge artifacts that would otherwise arise from zero-padding during Gaussian smoothing. Finally, the blurred output defines \(z_{-}\).

\paragraph{Gaussian Noise}
Given a batch input
\(
z_{+} \in \mathbb{R}^{B \times C \times H \times W},
\)
we compute a noise scale for each individual sample
\(
z_{+}^{(i)} \in \mathbb{R}^{C \times H \times W}.
\)

For each sample, we first measure its per-sample standard deviation:
\[
\sigma^{(i)} = \operatorname{std}\!\left(z_{+}^{(i)}\right),
\qquad
\tilde{\sigma}^{(i)} = \frac{\sigma^{(i)}}{\sigma_{\max}},
\]
where \(\sigma_{\max}\) is the maximum standard deviation observed within the batch.

We then define the noise magnitude as
\[
\gamma^{(i)} = \text{base\_scale}\,\bigl(1 - \tilde{\sigma}^{(i)}\bigr),
\]
where \text{base\_scale} is set to 255 in image space and to 1 in both latent and velocity spaces.

Finally, we inject Gaussian noise into each sample to get $z_{-}^{(i)}$:
\[
z_{-}^{(i)}
=
z_{+}^{(i)}
+
\gamma^{(i)}\,\varepsilon^{(i)},
\qquad
\varepsilon^{(i)} \sim \mathcal{N}(0, 1).
\]

\paragraph{Color Jitter}
Given a batch 
\(
z_{+} \in \mathbb{R}^{B \times C \times H \times W},
\)
we apply per-sample color jitter composed of brightness, contrast, and saturation adjustments.  
We first normalize the input to obtain \(z'\), and independently sample the jitter factors from
\[
\lambda_{\mathrm{b}},\,
\lambda_{\mathrm{c}},\,
\lambda_{\mathrm{s}}
\sim \mathcal{U}(1-\delta,\,1+\delta),
\qquad
\delta = 0.2.
\]

\textbf{Brightness.}
\[
z' \leftarrow \lambda_{\mathrm{b}}\, z_{+}.
\]

\textbf{Contrast.}
Let \(\mu = \mathrm{mean}(z')\) denote the global mean intensity:
\[
z' \leftarrow (z' - \mu)\lambda_{\mathrm{c}} + \mu.
\]

\textbf{Saturation.}
Let \(g = \mathrm{mean}_{c}(z')\) be the per-pixel channel average:
\[
z' \leftarrow (z' - g)\lambda_{\mathrm{s}} + g.
\]

These three operators are applied in a random order.  
The final result is clamped to \([0,1]\) and rescaled as needed to obtain \(z_{-}\).

\section{More Results}

In this section, we provide additional quantitative and qualitative results.

\subsection{ImageNet-1K Results with ODE Sampling}

Tab.~\ref{tab:imagenet-ode} reports the ImageNet-1K results under ODE sampling. Across all SiT backbones, integrating VeCoR yields clear improvements in the main quality metrics, achieving lower FID and higher IS under the same sampling budget (50 NFEs, Heun2).

Although FID, IS, and Precision improve, we observe mild decreases in sFID and Recall under certain configurations. A possible explanation is that, in a fully deterministic ODE setting, the additional signals from VeCoR about “where not to go’’ may guide the trajectory to remain closer to certain regions of the manifold, which could slightly limit the diversity of viable generation paths.

Overall, these shifts are small relative to the overall gains, and VeCoR remains beneficial under both SDE- and ODE-based sampling.

\begin{table}[t]
    \centering
\caption{Results on ImageNet-1K 256$\times$256 using SiT backbones (same seed, 50 NFEs, Heun2).}

   \resizebox{0.9\linewidth}{!}{
\renewcommand{\arraystretch}{1.2}
\begin{tabular}{lccccc}
\toprule
\textbf{Model} & \textbf{FID}  $\downarrow$  & \textbf{IS}  $\uparrow$ & \textbf{sFID}  $\downarrow$  & \textbf{Prec.}  $\uparrow$ & \textbf{Rec.}  $\uparrow$ \\
\midrule
SiT-S/2~\cite{ma2024sitexploringflowdiffusionbased} & 59.28 & 23.43 & 9.33 & 0.40 & \textbf{0.59} \\
+\textbf{VeCoR (Ours)} & \textbf{55.11} & \textbf{24.34} & \textbf{8.53} & \textbf{0.41} & \textbf{0.59} \\
\hline \addlinespace
SiT-B/2~\cite{ma2024sitexploringflowdiffusionbased} & 37.07 & 40.27 & \textbf{6.79} & 0.51 & \textbf{0.65} \\
+\textbf{VeCoR (Ours)} & \textbf{33.76} & \textbf{41.22} & 7.76 & \textbf{0.53} & 0.63 \\
\hline \addlinespace
SiT-L/2~\cite{ma2024sitexploringflowdiffusionbased} & 21.9 & 63.84 & \textbf{5.58} & 0.61 & \textbf{0.64} \\
+\textbf{VeCoR (Ours)} & \textbf{19.81} & \textbf{66.16} & 7.41 & \textbf{0.63} & 0.62 \\
\hline \addlinespace
SiT-XL/2~\cite{ma2024sitexploringflowdiffusionbased} & 18.59 & 72.05 & \textbf{5.30} & 0.63 & \textbf{0.64} \\
+\textbf{VeCoR (Ours)} & \textbf{16.55} & \textbf{76.17} & 7.21 & \textbf{0.65} & 0.62 \\
\bottomrule
\end{tabular}}

\label{tab:imagenet-ode}
\end{table}
\begin{figure*}[t]
    \centering
    \includegraphics[width=0.95\linewidth]{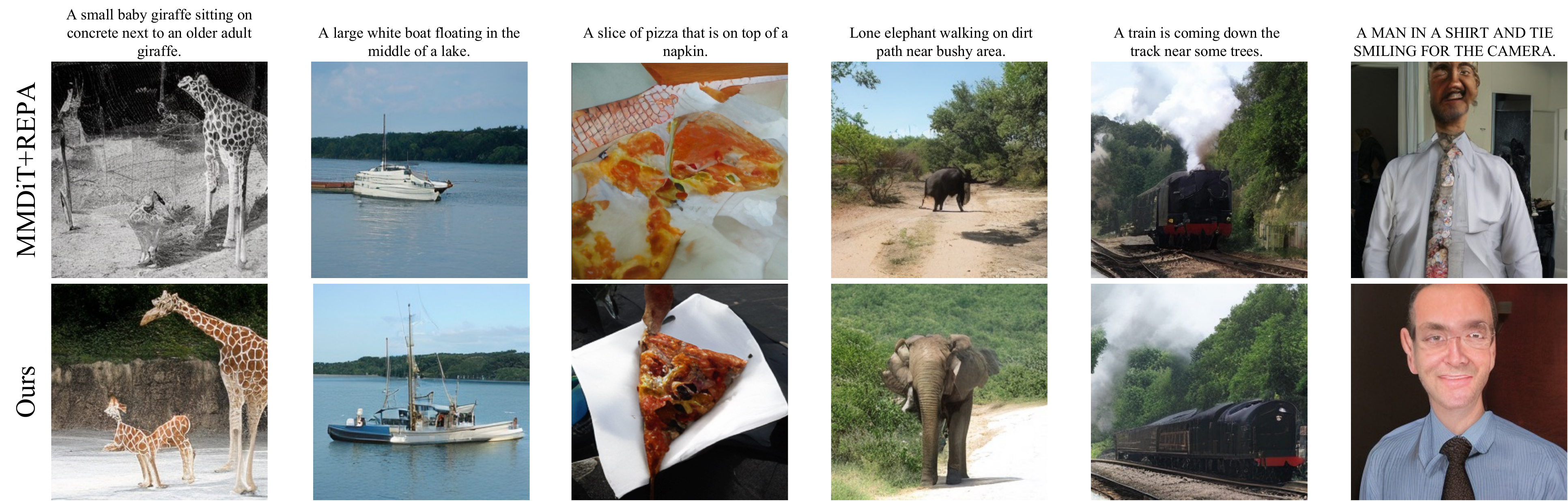}
    \caption{\textbf{Qualitative comparison on text-to-image generation (MS-COCO).} We use classifier-free guidance with $w = 2.0$ and using (same seed, 50 NFEs, Euler–Maruyama).}
    \label{fig:supple}
    \vspace{-5px}
\end{figure*}
\subsection{Text-to-Image Qualitative Results}

We provide the text-to-image visual comparisons in Fig.~\ref{fig:supple}, which illustrate that, under identical sampling conditions, incorporating VeCoR leads to outputs with better color consistency and stronger semantic alignment to the input prompts.

\section{On the Effectiveness of the VeCoR Loss}
For completeness, we provide the analytical form of our velocity contrastive regularization (VeCoR). 
Although its structure resembles the contrastive FM objective in $\Delta$FM~\cite{stoica2025contrastive}, the intent is fundamentally different: 
$\Delta$FM leverages contrastive signals \emph{across conditions} to enforce class-level separability, whereas our formulation applies contrastive supervision directly at the \emph{dynamics level} to enhance trajectory stability and suppress off-manifold drift during sampling. 
Thus, despite superficial similarities, the contrastive role in VeCoR is intrinsically distinct.

We begin by expressing the VeCoR objective in its expectation form:
\begin{equation}
\label{eq:expectation_vecor}
\begin{aligned}
\mathcal{L}^{(\mathrm{VeCoR})}(\theta)
=
\mathbb{E}\Big[
&
\|\mathbf{v}_{\theta}(\hat{x}_t,t)-\hat{\mathbf{v}}_{+}\|_2^{2}
\\
&\;-\;
\lambda\sum_{k=1}^{K}
\|\mathbf{v}_{\theta}(\hat{x}_t,t)-\hat{\mathbf{v}}_{-}^{(k)}\|_2^{2}
\Big],
\end{aligned}
\end{equation}
where the expectation is taken over timesteps, perturbed states $\hat{x}_t$, and injected noise.

\vspace{0.5em}
\noindent\textbf{Step 1: Expand and collect quadratic terms.}
Let $\mathbf{v}_\theta=\mathbf{v}_\theta(\hat{x}_t,t)$ for brevity.  
Expanding the squared terms and applying linearity of expectation yields
\begin{equation}
\label{eq:expanded_form}
\begin{aligned}
\mathcal{L}^{(\mathrm{VeCoR})}(\theta)
=
\mathbb{E}\Big[
&(1-\lambda K)\,\mathbf{v}_{\theta}^{\top}\mathbf{v}_{\theta}
\\
&-2\,\mathbf{v}_{\theta}^{\top}
\Big(
\hat{\mathbf{v}}_{+}
-
\lambda\sum_{k=1}^{K}\hat{\mathbf{v}}_{-}^{(k)}
\Big)
\Big]
+\text{const},
\end{aligned}
\end{equation}
where the constant aggregates all terms independent of $\mathbf{v}_\theta$.

\vspace{0.5em}
\noindent\textbf{Step 2: Compute the minimizer.}
Taking the gradient of \eqref{eq:expanded_form} with respect to $\mathbf{v}_{\theta}$ and setting it to zero gives
\begin{equation}
2(1-\lambda K)\,\mathbf{v}_{\theta}^{*}
=
2\,\mathbb{E}
\Big[
\hat{\mathbf{v}}_{+}
-
\lambda\sum_{k=1}^{K}\hat{\mathbf{v}}_{-}^{(k)}
\Big].
\end{equation}

Dividing both sides by $2(1-\lambda K)$ yields the closed-form solution:
\begin{equation}
\label{eq:optimal_closed_form}
\mathbf{v}_{\theta}^{*}
=
\frac{
\mathbb{E}[\hat{\mathbf{v}}_{+}]
-
\lambda\sum_{k=1}^{K}\mathbb{E}[\hat{\mathbf{v}}_{-}^{(k)}]
}{
1-\lambda K}.
\end{equation}

This derivation shows that the VeCoR objective preserves the FM fixed point while adding a contrastive correction that suppresses destabilizing dynamical alternatives. 
Unlike $\Delta$FM---whose contrastive term separates flows across conditioning labels---the negative velocities in VeCoR represent dynamical directions that would drive trajectories toward undesirable, off-manifold evolution. 
In this view, VeCoR acts as a corrective force that steers the predicted velocity away from off-manifold directions and reinforces stable, data-consistent trajectories. 
To maintain this behavior in a mathematically well-posed manner, the quadratic coefficient $(1 - \lambda K)$ must remain positive, ensuring that the objective retains a proper minimization structure. 
This leads to the requirement
\[
\lambda K < 1,
\]
which prevents the loss from becoming ill-conditioned.

\end{document}